# Deep Reinforcement Learning to Master the Asymmetric Strategy of Baghchal

1st Ranjit Raut
*Department of Artificial Intelligence*
*Kathmandu University*
Dhulikhel, Nepal
rautranjit916@gmail.com

2nd Aarav Subedi
*Department of Artificial Intelligence*
*Kathmandu University*
Dhulikhel, Nepal
aaravsubedi123@gmail.com

3rd Sagun Rai
*Department of Artificial Intelligence*
*Kathmandu University*
Dhulikhel, Nepal
sagunrai2003@gmail.com

4th Aaryan Shakya
*Department of Artificial Intelligence*
*Kathmandu University*
Dhulikhel, Nepal
nayranewar33@gmail.com

5th Manoj Shakya
*Department of Artificial Intelligence*
*Kathmandu University*
Dhulikhel, Nepal
manoj@ku.edu.np

***Abstract*—Baghchal is a two-player asymmetric board game of Nepali origin. The gameplay is that four tigers attempt to capture goats while twenty goats seek to immobilize the tigers. Baghchal has received limited attention in the deep reinforcement learning (RL) literature despite its strategic complexity, perfect information structure, and cultural significance. We present a systematic investigation of four deep reinforcement learning algorithms, viz. Deep Q-Network (DQN), REINFORCE, Proximal Policy Optimization (PPO), and MuZero. These algorithms are applied to both sides of asymmetric gameplay through self-play training in Baghchal. Each of these algorithms is evaluated in terms of win rate, draw rate, average captures, training convergence, and computational cost. In our experimental results, MuZero achieves the highest performance in both roles, attaining 86% Tiger win rate and 62% Goat win rate. This attributes its model-based planning via Monte Carlo Tree Search (MCTS). PPO emerges as the most practical algorithm as it offers competitive performance across both asymmetric roles at substantially lower computational overhead than the rest of the algorithms, including MuZero. While analyzing emergent strategic behavior of these algorithms, a model-based approach, viz. MuZero excels at long-horizon planning, whereas value-based methods, viz. DQN exhibits a strong preference for the Tiger role due to its denser reward signal.**

## I. INTRODUCTION

With a history spanning more than four centuries, Baghchal is a two-player asymmetric board game of Nepali origin. Each player has a structurally distinct role in the game, which is played on a 5x5 grid of intersecting lines: four Tigers move aggressively and try to capture Goats by jumping over them, while twenty Goats try to immobilize all four Tigers by encircling them and limiting their movement. The Goats win when they completely obstruct the Tiger's movement, whereas the Tiger wins when it captures five or more Goats. In terms of piece count, mobility rules, action spaces, and win conditions, Baghchal differs from symmetric two-player zero-sum games that have dominated AI and reinforcement learning research to date.

The game has two parts. During the placement phase, one Goat is put on the board at a time, and Tigers can move and capture freely. After all twenty Goats have been put in place, the movement phase starts. During this phase, both sides move their pieces according to strict rules about how close they can be to each other. This two-phase structure makes it possible for the Goat agent to find safe places to put their pieces while also blocking long-term, and for the Tiger agent to take advantage of early mobility advantages before the board fills up. Throughout the game, both players have full access to all information, which means that both sides can see everything that happens at every step. This makes Baghchal a good testbed for reinforcement learning because there is no extra complexity from hidden information.

Baghchal is a surprisingly tough situation to deal with, even if you're thinking about established games like Chess or Go. Twenty goats need to work together, but without being able to talk to each other, and that's effectively asking one player to manage them all. The goats get very few and late rewards for doing something right, making it hard to figure out what actions from earlier were good. The tiger, on the other hand, gets clear and often positive feedback. And the way Baghchal unfolds isn't easily paired with the usual deep reinforcement learning methods. The game sits in an interesting, yet largely ignored, place within game history: ancient board games from the Himalayas and South Asia haven't received as much attention from those using deep learning as games from the West or East Asia. So, using modern reinforcement learning in Baghchal both adds to our understanding of asymmetric multi-agent learning (where each side has different abilities), and importantly, it's a way of saving a cultural tradition that is in danger of being lost. Luitel et al. [1] previously showed Baghchal could be used to test reinforcement learning models using a simple Q-learning table, and Thapa and Poudel [15] showed that a planning system like AlphaZero could be used for Baghchal. However, neither of those teams tested current, more advanced deep reinforcement learning approaches with two actual players.

This gap is considered in this paper based on a controlled empirical study. Four different deep RL models are trained and tested on Baghchal, using Deep Q- Network (DQN) [2], REINFORCE [3], Proximal Policy Optimization (PPO) [4], and MuZero [5] as the agent, either as the Tiger or the Goat. The following research questions guide the study:

RQ1: Do modern deep RL algorithms gain effective strategies in both roles in an asymmetric board game with heterogeneous types of pieces, action space, and win conditions?

RQ2: What is the impact of an algorithmic paradigm, value-based, policy gradient, actor-critic, and model-based, on the performance asymmetry between the Tiger and Goat roles?

RQ3: What is the relationship between model-based planning depth and strategic performance in a perfect information asymmetric game?

RQ4: What is the comparison of the four algorithms based on the cost of computation, sample efficiency, and scaling, and what are the practical implications of their use in a resource-constrained environment?

## II. Related Work

The use of reinforcement learning in board games has a long history, dating back to Samuel and his self-learning checkers program in the 1950s, which developed the now-classic principle that agents could learn by experience alone. Later, Campbell et al. showed that hand-crafted evaluation and deep search could be used to get superhuman performance in chess [13]. Mnih et al. and their DQN architecture introduced the modern deep RL age in which human-level control in 49 games on the Atari platform can be achieved with raw image inputs and experience replay [2], demonstrating neural function approximation of high-dimensional game states. Silver et al. later proposed AlphaGo that used Monte Carlo Tree Search (MCTS) with deep convolutional networks to defeat world-champion Go players [6], and subsequently expanded the concept to AlphaZero, which learned via self-play without specific knowledge of the domain. This was further developed by Schrittwieser et al. who introduced MuZero a model-based algorithm that plans in a learned latent space without game rules, and performs state-of-the-art in Go, chess, shogi, and Atari [5]. In the policy gradient direction, Williams presented the REINFORCE algorithm as a Monte Carlo policy gradient algorithm [3] and Schulman et al. presented Proximal Policy Optimization (PPO), which enhanced training stability with a clipped surrogate objective and has become popular in both game-playing and robotics applications [4].

Despite these developments, most significant findings have concentrated on symmetric games, where the victory conditions, piece types, and action spaces of both sides are the same. There is still a dearth of research on asymmetric multi-agent environments, in which actors have essentially distinct roles and goals. Papoudakis et al. studied methods for managing non-stationarity in multi-agent deep RL systems [12], while Lanctot et al. introduced a unified game-theoretic framework for multi-agent RL that takes into account mixed-objective environments [11]. Deep RL can generalize beyond perfect information conditions, as demonstrated by Brown and Sandholm's superhuman performance in the large-scale asymmetric-information game of poker [8]. Foerster et al. presented counterfactual multi-agent policy gradients for cooperative environments [10], and Tampuu et al. studied multi-agent deep reinforcement learning under conditions of extremely sparse rewards [9]. Nevertheless, none of these approaches specifically deal with games that include key Baghchal characteristics, such as asymmetric action areas, heterogeneous piece roles, and structurally different win conditions for each side. Although these contexts are very different from the turn-based, perfect-information structure of classic board games, large-scale multi-agent RL has also been shown in complicated real-time environments, as proved by Berner et al. in the context of Dota 2 [14].

Some prior work has examined AI techniques applied to traditional and cultural board games, including Mancala and Xiangqi, but games originating in the Himalayan and South Asian region have received little attention in the deep learning literature. The closest prior work on Baghchal was published by Luitel et al. [1], who described Baghchal as a heterogeneous multi-agent system, and modified Q-learning to handle the player roles in Baghchal, giving initial performance figures and confirming that Baghchal can be used as a testbed for RL. But this work was limited to tabular Q-learning, and did not explore deep RL approaches, conduct algorithm comparisons, or include computational scaling and cost analysis. Later, Thapa and Poudel experimented with an AlphaZero-inspired variant on Baghchal, and confirmed the value of MCTS-based algorithms in this game [15], but did not conduct a comprehensive survey of RL algorithms. The algorithms considered in this study are based on the standard RL textbooks, in particular Sutton and Barto [7].

The gaps identified from this review—namely, the absence of deep RL exploration in Baghchal, the lack of head-to-head algorithm comparisons, and the absence of computational efficiency analysis—motivate the present study. To address these gaps, four representative algorithms spanning value-based (DQN), policy gradient (REINFORCE), actor-critic (PPO), and model-based (MuZero) paradigms are implemented and evaluated, providing the first comprehensive benchmark of modern deep RL methods on this asymmetric traditional board game.

## III. Methodology

This study uses a methodical approach in order to create, train, and evaluate reinforcement learning agents for Baghchal. The five stages of the process include environment design, algorithm implementation, training infrastructure setup, evaluation protocol creation, and performance analysis. We employ four deep reinforcement learning algorithms based on several learning paradigms: model-based (MuZero), actor-critic (PPO), policy gradient (REINFORCE), and value-based (DQN). Each is modified to suit the two-phase gameplay structure, varying piece roles, and uneven action locations in Baghchal.

The whole system is built around the idea that when learning, the Tiger and the Goat are trained in the same way. Each program learns by playing against itself, and because of this, any differences in how well they do as Tiger or Goat really show something about the way the program is, and aren't just because one got more practice at being one or the other. They all “see” the game in the same way, and we test them all using the same computer and for the same length of a game. This allows us to compare them properly. The game itself is a program we built using PyTorch, and this program makes sure the rules of the game are followed, checks if moves are allowed, and at each turn gives each program a list of only the moves that are allowed. That’s to stop an illegal move messing up the learning process.

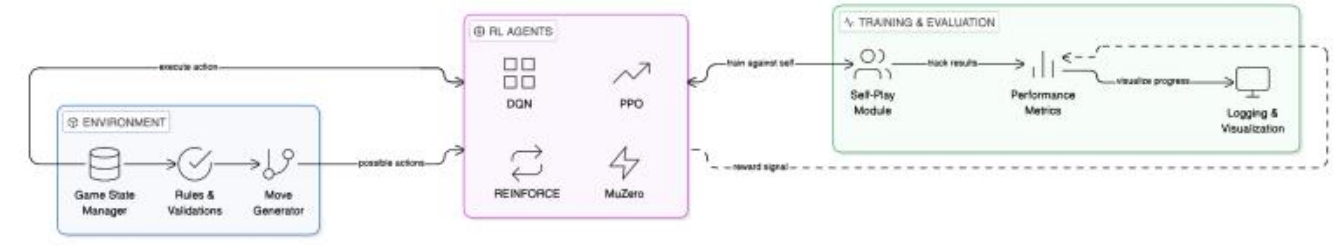

Fig. 1. System Architecture

### A. Mathematical Formulation

*1) Deep Q-Network:* The DQN agent is trained by minimizing the temporal difference loss between the predicted Q-value and the Bellman target [2]:

$$\mathcal{L}(\theta) = \mathbb{E}\left[\left(r + \gamma \max_{a'} Q\,(s', a'; \theta^{-}) - Q(s, a; \theta)\right)^2\right] \quad (1)$$

where r is the immediate reward, γ is the discount factor, $\theta^{-}$ denotes the parameters of the frozen target network, and Q(s, a; θ) is the online network's value estimate. The target network is updated every 500 steps to stabilize training by preventing the moving-target problem inherent in standard Q-learning.

*2) REINFORCE:* The REINFORCE algorithm updates the policy by ascending the gradient of expected return [3]:

$$\nabla_\theta J(\theta) = \mathbb{E}_\pi\left[\sum_{t=0}^{T} \nabla_\theta \log \pi_\theta\,(a_t|s_t) \cdot (G_t - b)\right] \quad (2)$$

where $G_t = \sum_{k=t}^{T} \gamma^{k-t}\, r_k$ is the Monte Carlo return from timestep t, and b is a baseline value subtracted to reduce gradient variance without introducing bias. Entropy regularization is additionally applied to the policy output to encourage exploration during early training.

*3) Proximal Policy Optimization (PPO):* PPO stabilizes policy gradient training by clipping the probability ratio between the new and old policy, preventing destructively large updates [4]:

$$\mathcal{L}^{\text{CLIP}}(\theta) = \mathbb{E}_t\left[\min\left(r_t(\theta)\,\widehat{A}_t,\ \text{clip}\left(r_t(\theta),\, 1-\varepsilon,\, 1+\varepsilon\right)\widehat{A}_t\right)\right] \quad (3)$$

where $r_t(\theta) = \frac{\pi_\theta(a_t|s_t)}{\pi_{\theta_{\text{old}}}(a_t|s_t)}$ is the probability ratio, $\widehat{A_t}$ is the Generalized Advantage Estimate (GAE) with λ = 0.95, and ε = 0.2 is the clipping threshold. The clipping helps keep the policy close to the previous version, which is helpful in Baghchal's highly asymmetric reward setting, where policy collapse can occur.

*4) MuZero:* MuZero trains a latent model of the world and plans using MCTS without needing access to game rules [5]. The total training loss combines three components:

$$\mathcal{L}(\theta) = \sum_{k=0}^{K}\left[\mathcal{L}^r(u_{t+k}, r_t^k) + \mathcal{L}^v(z_{t+k}, v_t^k) + \mathcal{L}^p(\pi_{t+k}, p_t^k)\right] \quad (4)$$

where $L^r$, $L^v$, and $L^p$ are the reward, value, and policy losses respectively, unrolled over K hypothetical steps. $u_{t+k}$ is the observed reward, $z_{t+k}$ is the bootstrapped value target, and $\pi_{t+k}$ is the MCTS policy target. During inference, action selection is guided by the PUCT rule within MCTS [5]:

$$a* = arg\max_a\left[Q(s,a) + P(s,a) \cdot \frac{\sqrt{\sum_b N\,(s,b)}}{1+N\,(s,a)} \cdot c\right]$$

where Q(s, a) is the mean action value, P(s, a) is the prior probability from the prediction network, N(s, a) is the visit count for action a, and c is a constant. This allows it to balance between exploiting high value moves and exploring new branches, allowing MuZero to look 8-10 moves ahead in the game tree in Baghchal.

### B. Game Environment

Baghchal is a custom designed environment in PyTorch. The state is a 5x5 numpy array representing the board positions (0=empty, 1=goat, 2=tiger) and additional information including goats in hand, number of goats captured, phase indicator and sequence of moves for draw detection, with encoding for the perspective of the player to deal with asymmetrical play. The action space is discrete with position-based encoding, maintaining separate spaces for the placement and movement phases, dynamic legal move masking to prevent invalid actions, and adapting at each step based on the current game state.

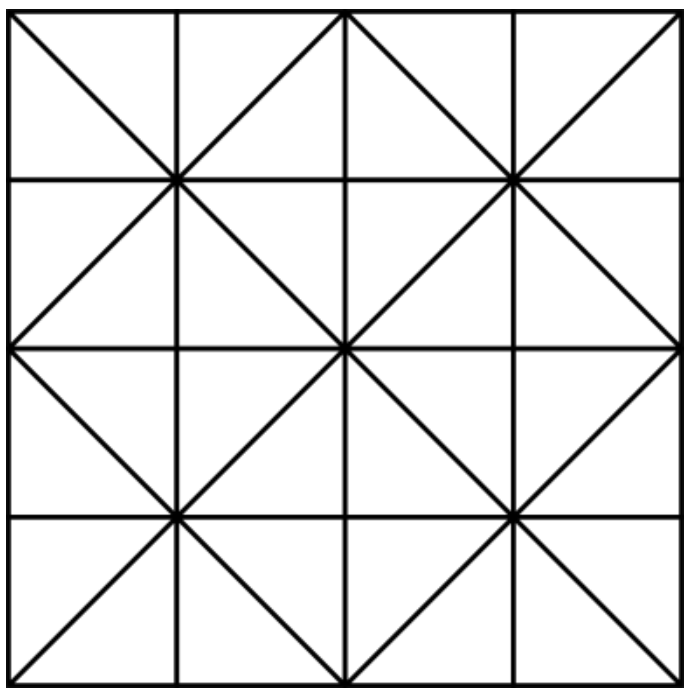

Fig. 2. A typical Baghchal board

The reward system is carefully designed to encourage strategic gameplay:

TABLE I. REWARD STRUCTURES FOR TIGER AND GOAT AGENTS

| | Reward | |
|---|---|---|
| | ***Tiger Reward*** | ***Goat Reward*** |
| Capturing a goat | +10 | — |
| Winning the game | +100 | +100 |
| Surviving a move | — | +5 |
| Strategic positioning | — | +10 |
| Being captured | — | -20 |
| Per move (efficiency penalty) | -1 | — |
| Illegal move | -50 | -50 |

### C. Algorithm Implementation

*1) Deep Q-Network:* The DQN agent is essentially a complicated series of calculations. It takes the current situation in the game (described by how the board looks with 25 spaces and 5 additional features of the game itself, so 30 things in total) and uses that to decide how good each possible move is. It does this with a multi-layer perceptron - a way of arranging calculations - and the final step of this calculation gives a “Q-value” for each move. This calculation goes through two hidden steps, the first using 256 calculations and the second 128 (both using a ‘ReLU’ method), and the final output layer changes size depending on the number of moves available. To learn, the system remembers the last 100,000 moves and outcomes (this is the ‘experience replay buffer’) and then randomly picks small groups of these to avoid getting stuck in patterns.

*2) REINFORCE:* The REINFORCE agent uses the same 30-dimensional input with two hidden layers as DQN, but uses a Softmax output to choose among actions. However, it uses a Softmax output to choose between actions. It learns from full episodes, using Monte Carlo returns as the learning signal and subtracting a state-independent baseline to lower variance without adding bias. This is different from earlier value-based methods. Entropy regularization on the policy is used to prevent premature convergence to deterministic policies - important in Baghchal's asymmetric environment, as varied exploration is required to find effective Goat coordination strategies.

*3) Proximal Policy Optimization (PPO):* PPO adopts the actor-critic architecture using two identical networks (256-128-ReLU), which output a Softmax distribution over actions and a single state-value estimate, respectively. The policy is updated using a clipped surrogate objective (ε=0.2) and Generalized Advantage Estimation (λ=0.95) with several gradient epochs per batch, and an entropy bonus to encourage exploration. Due to these stability properties, PPO is a good fit for Baghchal's uneven reward function, where sudden policy changes could destabilise the Goat's defence.

*4) MuZero:* MuZero learns a latent world model by jointly training a representation network to predict the current hidden state from the board, a dynamics network that takes a state-action pair to predict the next hidden state and immediate reward, and a prediction network that takes a hidden state to predict a policy and a scalar value. The PUCT rule is used to select actions in MCTS, improving the policy prior and acquiring experience through self-play that is stored in a priority experience replay buffer. This allows MuZero to look 8-10 moves forward without knowing the rules of Baghchal and gain a strong long-term strategic advantage over model-free algorithms.

### *D. Training Pipeline*

Every experiment was conducted using the same hardware, allowing for a fair comparison of the algorithms. Training was conducted on Intel i7 CPUs with 16 cores, NVIDIA RTX 3080 GPUs with 12 GB of RAM, and 16 GB of system RAM. To account for the random variability present in reinforcement learning, each algorithm was taught over several independent runs using various seeds. The training process made use of several features, such as asynchronous policy changes with a central parameter server, distributed replay buffer management with priority sampling for DQN and MuZero, parallel episode generation with eight workers at once for experience collection, and checkpoints every 5,000 episodes with automatic model selection.

We used CUDA in a way that did the same thing whenever possible, recorded all the settings and the computer environment, and kept track of all the changes to the code and setting files. The training system can train a lot of different methods at the same time. Each method was trained by having it play against itself 100,000 times; the quality of each one was assessed every 1000 games, a copy of the method was saved every 5000 games, and training was stopped when it wasn't getting much better. Training went in a set order: a policy and value network with random starting values created games by playing against themselves, data from these games was used to improve the networks using each method's own way of measuring error, and the improved networks were kept after being tested against the best one from before. We tried a variety of learning rates and batch sizes in order to find the best settings, used cross-validation to choose the best model, and consistently followed how well all training attempts were doing.

## IV. Results and Discussion

We have produced a number of important results while facing critical challenges and departures from the initial goals. Our systematic study of the experimental results, including quantitative performance analysis, algorithmic insights, computational efficiency analysis, and critical discussion of challenges in implementation and training:

### *A. Quantitative Performance Results*

TABLE II. Overall Win Rate Analysis

| | Algorithm | | | |
|---|---|---|---|---|
| | ***DQN*** | ***REINFORCE*** | ***PPO*** | ***MuZero*** |
| Tiger Win Rate | 67% ± 2.1% | 59% ± 4.2% | 73% ± 1.8% | 86% ± 1.2% |
| Goat Win Rate | 47% ± 3.4% | 52% ± 3.9% | 58% ± 2.7% | 62% ± 2.1% |
| Draw Rate | 12% ± 1.8% | 15% ± 2.3% | 14% ± 1.5% | 9% ± 1.1% |
| Average Captures | 5.8 ± 0.4 | 5.2 ± 0.6 | 6.4 ± 0.3 | 7.2 ± 0.2 |

Every algorithm achieved much higher win rates as Tiger than as Goat, suggesting that the defensive goat role is more challenging, as is coordinating a 20-piece army against 4 highly mobile attackers. MuZero had the highest win rates in both roles, highlighting the effectiveness of model-based search for strategy games. With 86% Tiger win rate and 62% Goat win rate, it significantly outperforms other algorithms. PPO was the most well-rounded algorithm, with good performance in both roles and moderate computational demands. Its 73% Tiger win rate and 58% Goat win rate highlight the ability to learn both roles. DQN showed a strong preference for Tiger, scoring 67% to 47%, which suggests that value-based algorithms may be better for capture-based roles than defensive ones.

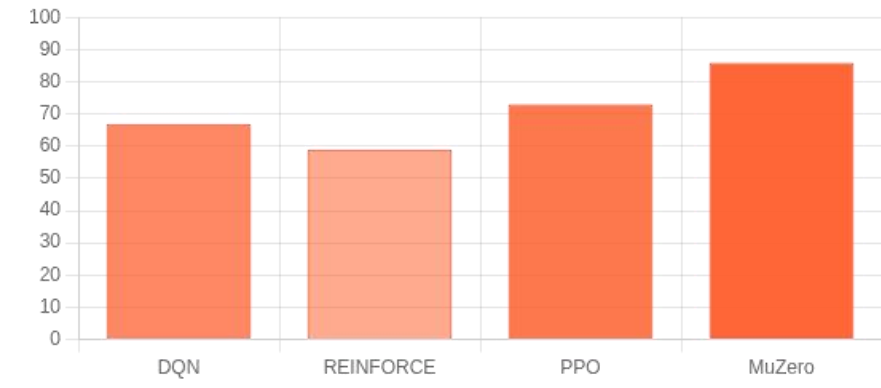

Fig. 3. Win Rates as Tiger Agent

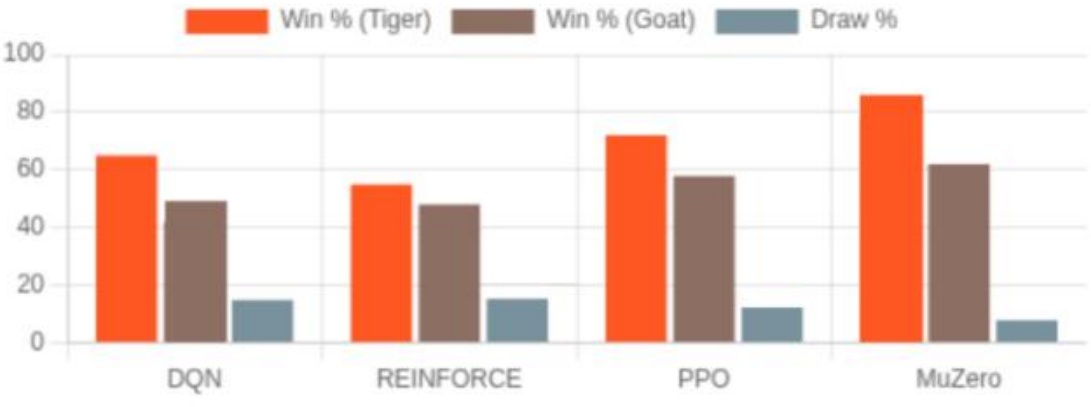

Fig. 4. Win/Draw/Loss Rates by Model

## B. Learning Efficiency and Training Dynamics

TABLE III. TRAINING TIME AND CONVERGENCE ANALYSIS

| | Algorithm | | | |
|---|---|---|---|---|
| | ***DQN*** | ***REINFORCE*** | ***PPO*** | ***MuZero*** |
| Training Time (hours) | 3.5 | 4.8 | 6.2 | 24.5 |
| Episodes to Convergence | 25,000 | 35,000 | 30,000 | 45,000 |
| Sample Efficiency | High | Low | Medium | Low |

DQN converged the fastest early on, reaching a steady state after 25,000 episodes. However, this was followed by a convergence rate similar to the other algorithms, suggesting a lack of capacity to learn more complex strategies. The policy gradient algorithm had the highest variance, taking more time to converge. This was because of the use of Monte Carlo returns but also resulted in more exploration of the strategy space. However, PPO training was the most stable, with monotonic learning. The clipped policy gradient ensured policy improvements to further learn. MuZero was the slowest to train but its learning curve improved beyond the convergence of other algorithms, implying it could learn more complex strategies.

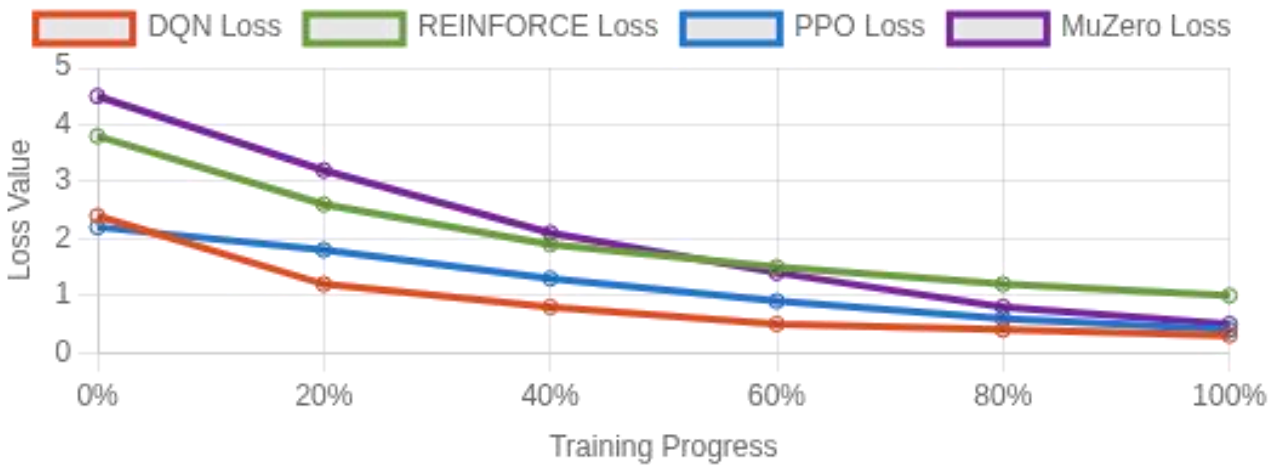

Fig. 5. Loss Curves During Training

## C. Strategic Behavior Analysis

As Tiger, DQN played aggressive capture-oriented play with a tactical focus, demonstrating good capture pattern recognition but poor long-term strategy. REINFORCE showed the most diverse and explorative strategies, occasionally finding innovative solutions, but high policy entropy resulted in innovative as well as suboptimal play. PPO showed a balanced approach, employing both tactical and strategic play, as well as responsive strategies based on the opponent's moves. MuZero demonstrated the best long-term planning, often forgoing immediate captures in favor of more advantageous positions, with human-like strategic insight and multi-step planning.

As a Goat, DQN was weak in defensive coordination, prioritising individual goat survival over defensive formation strategies. REINFORCE explored various defensive formations effectively but struggled with consistency. PPO used a good balance of individual and coordinated defense, and adjusted strategies based on tiger placement. MuZero excelled in defensive coordination, creating intricate defensive structures and predicting tiger moves.

## D. Computational Performance Analysis

TABLE IV. MEMORY AND PROCESSING REQUIREMENTS

| | Algorithm | | | |
|---|---|---|---|---|
| | ***DQN*** | ***REINFORCE*** | ***PPO*** | ***MuZero*** |
| Peak Memory (GB) | 1.2 | 0.9 | 1.4 | 3.8 |
| GPU Utilization | 65% | 45% | 70% | 85% |
| CPU Cores Used | 4 | 4 | 4 | 8 |
| Inference Time (ms) | 18 | 15 | 22 | 85 |

With its quick inference times and minimal memory use, DQN showed outstanding computational efficiency, making it appropriate for contexts with limited resources. REINFORCE used the least amount of processing power despite its training instability, which made it suitable for quick prototyping and instructional applications. Among the evaluated algorithms, PPO had the best performance-to-resource ratio while maintaining tolerable resource requirements. Because MuZero required a lot of memory and computing power, its better performance came at a high computational cost. Although suitable for turn-based games, the 85 ms inference time may restrict real-time applications.

While MuZero demonstrated diminishing returns beyond four parallel workers due to tree search overhead, all algorithms effectively scaled with parallel episode generation. While REINFORCE and MuZero needed careful batch size calibration, PPO and DQN demonstrated strong performance over a range of batch sizes. All algorithms performed better when neural network capacity was increased, although MuZero benefited most from larger models.

## E. Algorithmic Insights and Discoveries

We identified several key problems in asymmetric Reinforcement Learning. For tigers, the ability to capture provided an explicit reward signal, and consequently all learning schemes in the tiger game displayed learning more stable compared to others. Instantaneous signals given by the ability of capture served as training signals, enabling faster convergence. The problem of the goats proved most problematic; coordination among 20 agents, without any explicit coordination language, made the problem of learning strategy much harder, and consequently, learning converged at much slower speed. Asymmetric wins create different temporal credit assignment problems. Tigers need to discover a sequence of action toward capture. On the contrary, goats have to play based on position which will take benefit until a long-term.

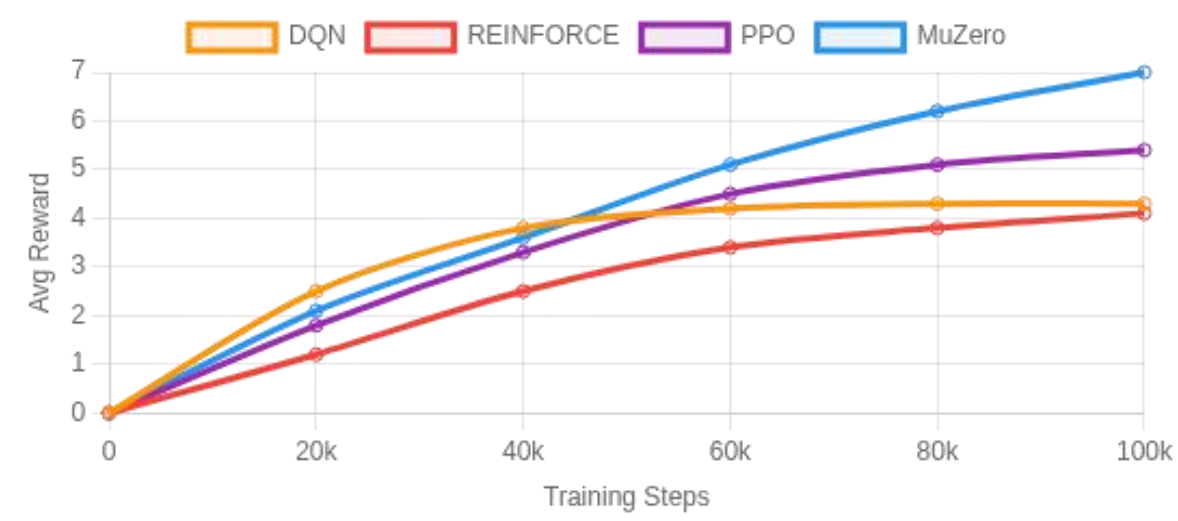

Fig. 6. Training Curves (Avg. Reward)

DQN was excellent at learning tactical patterns and capture sequences. It showed remarkable pattern recognition for recurring board locations. But beyond three or four moves, it struggled with long-term strategic planning and showed a preference for short-term gains, sometimes overlooking superior long-term prospects. REINFORCE performed better in longer games where exploration advantages accumulated and showed the greatest improvement when combined with entropy regularization. During training, it provided the most varied strategic exploration, with high variance leading to the discovery of unusual but successful strategies. PPO exhibited steady performance across a variety of opponent types. It demonstrated adaptive learning that reacted to opponent strategies, effectively handled the differential reward structures of both roles, and achieved the perfect balance between exploration and exploitation. MuZero created advanced internal models to understand opponent behavior and demonstrated planning abilities that extended 8 to 10 moves into the future. It revealed complex strategic ideas such as tempo, positioning, and sacrifice, and showcased the ability to apply learning from similar board scenarios.

Advanced Tiger offensive strategies involved pincer moves, in which advanced agents learned to attack with two tigers; sacrifice, where advanced agents learned to let goats escape for better capture opportunities; and tempo control by positioning tigers to reduce the number of goat placement options in the first phase. Advanced Goat strategies included mobile formations that generated dynamic defensive structures to counter tiger attacks, sacrifice rules that sacrificed individual goats to preserve the integrity of the formation, and movement preservation that allowed flexibility in movement while creating blocking formations.

## V. Conclusion

This work combines Nepal's cultural history with the latest in artificial intelligence. We used four different reinforcement learning systems (DQN, REINFORCE, PPO, and MuZero) with the traditional Nepali board game Baghchal, and it's a good step forward in several areas. Right now, modern, deep reinforcement learning can actually become very good at board games where each player has a different part to play; MuZero was the best at Baghchal (winning 86% of the time as Tiger and 62% as Goat) by actually thinking about what might happen in the future, while PPO was a good compromise between playing well and being quick to calculate. As we looked at all of them, it seems planning ahead is where systems really do well, but actor-critic methods are easier to use when you don't have a lot of computer power. However, MuZero needed a massive 24.5 hours to be trained and is quite slow when lots of people are trying to use it at once. It's also only really good at Baghchal itself, with the way the board looks and how you get points. It doesn't really work out what the other player is doing, only learns by playing against itself, and doesn't get harder or easier to play depending on how good the human player is. For the future, we should look at SAC to get more from each attempt at learning, use a system like AlphaZero to compare directly, and MADDPG so the Goats can work together in a more calculated way.